# Supervised Topical Key Phrase Extraction of News Stories using Crowdsourcing, Light Filtering and Co-reference Normalization

Luís Marujo[1,2], Anatole Gershman[1], Jaime Carbonell[1], Robert Frederking[1], João P. Neto[2]

[1] LTI/CMU, USA
[2] INESC-IST, Portugal

Luis.Marujo@inesc-id.pt, anatoleg@cs.cmu.edu, jgc@cs.cmu.edu, ref@cs.cmu.edu, Joao.Neto@inesc-id.pt

**Abstract**

Fast and effective automated indexing is critical for search and personalized services. Key phrases that consist of one or more words and represent the main concepts of the document are often used for the purpose of indexing. In this paper, we investigate the use of additional semantic features and pre-processing steps to improve automatic key phrase extraction. These features include the use of signal words and freebase categories. Some of these features lead to significant improvements in the accuracy of the results. We also experimented with 2 forms of document pre-processing that we call light filtering and co-reference normalization. Light filtering removes sentences from the document, which are judged peripheral to its main content. Co-reference normalization unifies several written forms of the same named entity into a unique form. We also needed a "Gold Standard" – a set of labeled documents for training and evaluation. While the subjective nature of key phrase selection precludes a true "Gold Standard", we used Amazon's Mechanical Turk service to obtain a useful approximation. Our data indicates that the biggest improvements in performance were due to shallow semantic features, news categories, and rhetorical signals (nDCG 78.47% vs. 68.93%). The inclusion of deeper semantic features such as Freebase sub-categories was not beneficial by itself, but in combination with pre-processing, did cause slight improvements in the nDCG scores.

**Keywords:** Automatic Key Phrase Extraction, Semantic Features, Pre-Processing

## 1. Introduction

In the last decade, the news consumption paradigm shifted from the traditional physical newspapers to personalized online news aggregation systems, such as News360, Google News, and Yahoo! News. These systems collect large amounts of news from various sources and provide an aggregate view of news on their websites and mobile applications. Fast and effective automated indexing is a critical problem for such services. Key phrases that consist of one or more words and represent the main concepts of the document are often used for the purpose of indexing. The precision and F1 measure of current state of the art automatic key-phrase extraction systems (AKE) is in the 30-50% range (Marujo et al., 2011; Medelyan et al., 2010; Witten et al., 1999). This makes improvements in AKE an urgent problem. In this work, we followed a fairly traditional approach of training a classifier to select an ordered list of the most likely candidates for key phrases in a given document. The main novelty of the paper is the use of additional semantic features and pre-processing steps. We tested several features, which to the best of our knowledge, have not been used for this purpose. These features include the use of signal words, freebase categories, etc. Some of these features lead to significant improvements in the accuracy of the results. We also experimented with 2 forms of document pre-processing that we call light filtering and co-reference normalization. Light filtering removes sentences from the document, which are judged peripheral to its main content. Co-reference normalization unifies several written forms of the same named entity into a unique form. In our experiments, both light filtering and co-reference normalization lead to small but noticeable improvements in the resulting accuracy of key phrase extraction. We also needed a set of "Gold Standard" (GS) labeled documents for training and evaluation. We used Amazon's Mechanical Turk[1] (Mturk) service to obtain these.

In this paper, we report our experiments with crowdsourcing for key phrase extraction and the results of our experiments with 2 new pre-processing steps and new features.

This paper is organized as follows: Section 2 presents the pre-processing steps; the description of the new features explored is presented in Section 3; the creation of a GS dataset using crowd-sourcing is described in Section 4; Section 5 details how the experiments were performed and their results, and Section 6 contains conclusions and suggestions for future work.

## 2. Pre-Processing

**Light Filtering:** our previous experiment with Portuguese-language broadcast news indicated that the elimination of about 10% of low-relevance sentences from the body of a news transcript results in a 2% improvement in AKE precision and recall. We hypothesized that similar improvements may be achieved in English-language news articles. We call this process *light filtering*. It is based on assigning a measure of relevance to each sentence of the article using *centrality-as-relevance* methods (Ribeiro et al. 2011). Centrality-as-relevance calculates pair-wise distances between sentences and finds a centroid for the article. The K sentences closest to the centroid are called the support set (SS). The distance between a sentence and the support set is used as a measure of this sentence relevance. Based on our previous experiments, we used 5 support sentences per document and removed the 10% of the most distant sentences from all documents using the Euclidean distance ($x$ and $y$ are vectorial sentence representation and $n$ designates the sentence length in words of the longest sentence):

$$D_{euclidean} = |x - y| = \sqrt{\sum_{i=1}^{n} |x_i - y_i|^2}$$

---

[1] https://www.mturk.com/

**Co-reference Normalization:** for stylistic reasons, journalists often use different forms of reference to the same named entities. For example, they might refer to Michael Jackson as Jackson or Michael. We hypothesized that normalizing such references would improve the AKE performance. We used ENCORE (Shah et al. 2011), a semi-supervised, ensemble co-reference resolution system to identify multiple forms of the same named entity and to normalize them into a single form (e.g., Michael Jackson).

## 3. Features

Typically, classifier-based Automatic Key-phrase Extraction systems tools include such features as TF-IDF, (Salton et al. 1975):

$$TF - IDF(t, D) = tf(t, d) \times idf(t, D)$$

$$idf(t, D) = log \frac{|D|}{1 + |\{d \in D: t \in d\}|}$$

where,
- tf(t,d) is the number of occurrences of term or phrase t in document d;
- |D| is the number of documents in the corpus
- $|\{d \in D: t \in d\}|$ is the number of documents containing term or phrase t

Other features use position on the page (Witten et al., 1999), number of words in the phrase (Medelyan et al., 2011), part of speech tags (Marujo et al., 2010), etc. We decided to test two additional kinds of features: semantic and rhetorical. We used three levels of semantic features – shallow semantic features, top-categories and sub-categories. The **shallow semantic** features consist of five dimensions:
1. the number of characters in a phrase - empirically noun words that are long tend to be relevant,
2. the number of named entities - very often named entities are important key phrases; typically this number is 0, 1, or 2,
3. the number of capital letters - the identification of acronyms is the main reason to include this feature,
4. the Part-of-Speech (POS) pattern of the phrase (e.g., <noun>, <adj, noun>, <adj, adj, noun>, etc.) – noun and noun phrases are the most common pattern observed in key phrases, verb and verb phrases are less frequent, and key phrases made of the remaining POS tags are rare;
5. we assign a distinct integer to each pattern,
6. the frequency of the phrase in the LDC HUB4 dataset[2] - to be precise we use the corresponding entry of 4-ngram model created using the dataset. The model was compressed using the Minimal Perfect Hash method (Guthrie et al., 2010) to reduce both memory consumption and access times to the model. We used smooth-nlp toolkit[3] to compress the model.

The **top-categories** we used are: *Technology, Crime, Sports, Health, Art and Culture, Fashion, Science, Business, World Politics*, and *U.S. Politics*. We also used 85 **sub-categories** taken from the Freebase domain names[4]. These included *American Football, Baseball, Book, Exhibitions, Education Engineering, Music, etc*. Both the top-categories and the sub-categories are used as binary features of a phrase. The top-category of each phrase is obtained from the document source category and the sub-categories are extracted by looking up the phrase in a Freebase dump.

Authors of news articles use various **rhetorical devices** to direct the reader's attention. The following eleven types of signals have been identified in the literature (Fry et al., 1990):
1. Continuation - there are more ideas to come, e.g.: *moreover, furthermore, in addition, another.*
2. Change of direction – there is a change of topic, e.g.: *in spite of, nevertheless, the opposite, on the contrary.*
3. Sequence – there is an order in the presenting ideas, e.g.: *in first place, next, into (far into the night).*
4. Illustration – gives an example, e.g.: *to illustrate, in the same way as, for instance, for example.*
5. Emphasis – increases the relevance of an idea, these are the most important signals, e.g.: *it all boils down to, the most substantial issue, should be noted, the crux of the matter, more than anything else.*
6. Cause, Condition, or result – there is a conditional or modification coming to following idea, e.g.: *if, because, resulting from.*
7. Spatial signals – denote locations, e.g.: *in front of, between, adjacent, west, east, north, south, beyond.*
8. Comparison/contrast – comparison of 2 ideas, e.g.: *analogous to, better, less than, less, like, either.*
9. Conclusion – ending the introduction of the idea and may have special importance, e.g.: *in summary, from this we see, last of all, hence, finally.*
10. Fuzz – there is an idea that is not clear, e.g.: *looks like, seems like, alleged, maybe, probably, sort of.*
11. Non-word emphasis, e.g.: exclamation point (*!*), "quotation marks".

---

[2] http://www.ldc.upenn.edu/Catalog/CatalogEntry.jsp?catalogId=LDC2000S88
[3] http://tinyurl.com/MphfCompres
[4] http://www.freebase.com/schema

Figure 1: Example of Amazon Mechanical Turk HIT used for creating the "Gold Standard"

We hypothesized that sentences containing such signals are more likely to contain key phrases. We used each of these eleven types of signals as a feature of a phrase. The values are the number of signals in the containing sentence.

## 4. Crowdsourcing

To evaluate our hypotheses, we needed a set of news stories with the corresponding key phrases. Obtaining such a set presents both conceptual and practical difficulties. Designations of key phrases are subjective decisions of each reader with relatively little agreement among them. Our solution was to use multiple annotators for the same news story and assign each phrase a score equal to the number of annotators who selected this phrase as a key phrase. We then ordered the phrases based on these scores and kept only the phrases selected by at least 90% of the annotators. We used Amazon's Mechanical Turk service to recruit and manage our annotators. To the best of our knowledge, this has not been done before for this purpose. Each assignment (called HIT) consisted of clicking on the most meaningful sequences of words in a news story.

We provided several examples shown on Figure 1. Annotating one story was a HIT and it paid $0.02 if accepted. We selected 50 stories for each of the 10 categories and created 20 HITs for each of the 500 stories in our set. An individual performer could only do one HIT per story. Unfortunately, this creates a practical problem of uneven quality of performers: some of the performers used bad shortcuts to do a HIT, producing meaningless results. We used several heuristics to weed out bad HITs. For example, the inclusion of stop words, very long sequences (> 10 words), and very fast work completion (< 30 seconds) usually indicated a bad HIT. As a result, we were able to keep 90% of HITs for each story. We created a "Gold Standard" set of 500 annotated news stories. The average number of key phrases per story was about 40 (39.72 to be exact). This number includes all of the key phrases occurring in all good Hits. However, the average agreement between workers was only 55% (10 workers).

The Gold Standard was split into two sets: 450 stories for training and 50 for testing.

## 5. Experiments and Results

For our experiments, as a baseline, we used the (Medelyan et al., 2010) – a state-of-the-art supervised key phrase extractor based on a bagging[5] over C4.5 decision tree classifier (Breiman et al., 1996; Quinlan, 1994). The shallow semantics features were previously used by us for Portuguese news stories (Marujo et al., 2011). In this work, we needed to adapt it to English. We used the MorphoAdorner name recognizer[6] for Named Entities Recognition and the Stanford Log-linear POS tagger (Toutanova et al., 2003). The frequency of the key phrase was computed using a 4-gram domain model - about 62K unigrams, 11.000M bigrams, 5.700M trigrams, and 4.000M 4-grams generated from LDC HUB4 dataset[7]. In the table below, it is the second testing condition (after the baseline) - we call it SS for Shallow Semantics. The third testing condition (TC) was the inclusion of the 10 top-level semantic categories. The fourth was the inclusion of the rhetorical signals (RS) and the fifth was the inclusion of the Freebase sub-categories (SC). We also tested the system with and without pre-processing - Light Filtering (LF) and Co-reference Normalization (CN).

---

[5] Bagging is a machine learning meta-algorithm, which is used with many classification tecnhiques, being very effective with decision tree models by reducing the variance associated with the predictions, by that means improving the result.

[6] http://morphadorner.northwestern.edu/

[7] http://www.ldc.upenn.edu/Catalog/CatalogEntry.jsp?catalogId=LDC2000S88

| Condition | nDCG | Precision |
|---|---|---|
| Baseline | 68.93% | 49.4% |
| Baseline + SS | 76.29% | 55.0% |
| Baseline + SS + TC | 77.05% | 50.8% |
| Baseline + SS + TC + RS | 78.47% | **56.2%** |
| Baseline + SS + TC + RS + SC | 75.45% | 53.4% |
| Baseline + SS + TC + RS + SC + CN | 77.87% | 54.8% |
| Baseline + SS + TC + RS + CN + LF | 77.77% | 53.8% |
| Baseline + SS + TC + RS + SC + CN + LF | **78.99%** | 55.4% |

Table 1: Results of our AKE system when extracting 10 key phrases (p-value < 0.05)
Legend: SS – Shallow Semantics
TC – Top Categories
RS – Rhetorical Signals
SC – Sub-Categories from Freebase
CN – Co-reference Normalization pre-processing
LF – Light Filtering pre-processing

In our experiments, we limited the number of extracted key phrases to 10. This made the calculation of recall meaningless. Consequently, we used two measures to evaluate the results: precision (P) and nDCG (Jarvelin et al., 2000) (normalized Discounted Cumulative Gain). To calculate precision, we first define True Positives (TP) as key-phrases co-occurring in both Gold Standard and AKE system list results. False Positives (FP) are phrases mistakenly identified as key-phrases by our system.

$$P = \frac{TP}{TP + FP}$$

To use the latter metric (nDCG), we assigned a score to each key phrase in a story. The score or relevance is equal to the number of human annotators who selected this phrase. We used these scores to sort the key phrases in a monotonically decreasing order of relevance.

$$DCG = rel_1 + \sum_{i=2}^{n} \frac{rel_i}{log_2 i}$$

Where $rel_i$ represents the relevance score of each key phrase at rank i, i.e., the number of workers that selected a phrase as relevant. For normalizing DCG, an ideal ordering for the list of key phrases is needed. The DCG of this ideal ordering is iDCG:

$$nDCG = \frac{DCG}{iDCG}$$

Table 1 presents the AKE results with new features and pre-processing steps. The baseline corresponds to the Maui standard system, without the Wikipedia based features because they did not improve the results in our preliminary experiments

## 6. Discussion and Conclusions

Our data indicates that the biggest improvements in performance were due to the shallow semantics features, top news categories and rhetorical signals (nDCG 78.47% vs. 68.93%). The inclusion of Freebase sub-categories was not beneficial by itself but in combination with pre-processing it did cause slight improvements in the nDCG scores.

It is interesting to compare our results with human performance. Since human annotators did not order their key phrases, we randomly ordered them 100 times for each annotator and computed the average nDCG score against the gold standard. The result was 64.63%, which is considerably lower than the system's performance. This may be due to the relative lack of agreement among human annotators and sorting. Since the system is trained on the intersection of phrases (90% agreement among annotators), it seems to produce better results when measured against the weighted ordering.

While the accuracy of automatic key phrase extraction may never be very high, it may be sufficient to improve the accuracy of news clustering by boosting the weights of more significant words and phrases as compared to the traditional TF-IDF scores. We hope to verify this fact in future work.

## 7. Acknowledgements

We thank the anonymous reviewers for their helpful comments.

Support for this research by FCT through the Carnegie Mellon Portugal Program and under FCT grant SFRH/BD/33769/2009.

## 8. References

Breiman, L., (1996). Bagging predictors. Machine Learning, 24(2), pp. 123--140.

Guthrie, D.; Hepple, M. and W. Liu, (2010). Efficient minimal perfect hash language models. *In Proceedings of LREC'10*. Valletta, Malta. European Language Resources Association (ELRA).

Fry, E.B., Polk, J.K., and Fountoukidis, D. *The New Reading Teacher's Book of Lists*. Prentice Hall, 1990.

Jarvelin, K.; Kekalainen, J. (1996). IR evaluation methods for retrieving highly relevant documents. *Proceedings of the 23rd annual international ACM SIGIR*, pp. 41-48.

Marujo, L.; Viveiros, M.; Neto, J. P. (2011). Keyphrase Cloud Generation of Broadcast News. *In proceeding of 12th Annual Conference of the International Speech Communication Association (Interspeech 2011)*.

Medelyan, O., Perrone, V., and Witten, I.H. Subject metadata support powered by Maui. *Proceedings of the 10th annual joint conference on Digital libraries - JCDL '10*, ACM Press (2010), 407.

Quinlan, J. (1994). C4.5: programs for machine learning. Machine Learning, 16, pp. 235--240.

Ribeiro, R.; Matos, D. M. (2011) Revisiting Centrality-as-Relevance : Support Sets and Similarity as Geometric Proximity. *Journal of Artificial Intelligence Research*

Salton, G.; Yang, C.S (1975). A theory of term importance in automatic text analysis. *Journal of the American society, 26* (1). pp. 33—44

Witten, I.H.; Paynter, G.W.; Frank, E.; Gutwin, C.; Nevill-Manning, C.G. (1999), KEA: Practical automatic keyphrase extraction. *Proceedings of the fourth ACM conference on Digital libraries*, ACM pp. 254--255.

Shah, R.; Lin, B.; Rosa, K.D.; Gershman, A.; Frederking, R. (2011). Improving Cross-Document Co-Reference with Semi-Supervised Information Extraction Models. *Proceedings of the Symposium on Machine Learning in Speech and Language Processing (MLSLP 2011)*,

Toutanova, K.; Klein, D.; Manning, C.; Singer. Y. (2003). Feature-Rich Part-of-Speech Tagging with a Cyclic Dependency Network. In *Proceedings of HLT-NAACL 2003*, pp. 252-259.